\documentclass[preprint,11pt,3p,authoryear]{elsarticle}

\usepackage{fullpage}
\usepackage{wrapfig}
\usepackage{comment}
\usepackage[utf8]{inputenc} 
\usepackage[T1]{fontenc}    
\usepackage{url}            
\usepackage{booktabs}       
\usepackage{siunitx} 
\usepackage{nicefrac}       
\usepackage{microtype}      
\usepackage{xcolor}         
\definecolor{bestcolor}{rgb}{1,0,0} 
\usepackage{subcaption}
\usepackage{caption}
\usepackage[normalem]{ulem}
\usepackage{tikz}
\usepackage{comment}

\usepackage{amsfonts}       
\usepackage{amssymb}
\usepackage{amsmath}
\usepackage{amsthm}
\usepackage{mathtools}
\usepackage{multicol}
\usepackage{multirow}

\usepackage{color}

\usepackage{array}
\usepackage{algorithm}
\usepackage{algpseudocode}
\usepackage[shortlabels]{enumitem}

\usepackage{longtable}
\usepackage{booktabs}
\usepackage{geometry}
\usepackage{subcaption}
\graphicspath{{./Figures/}} 
\usepackage[normalem]{ulem}
\newcounter{ToDo}
\newcounter{gaocomm} 
\newcounter{Note}
\definecolor{blue-violet}{rgb}{0.00,0.75,0.90}
\definecolor{mygreen}{rgb}{0.0, 0.5, 0.0}
\definecolor{awesome}{rgb}{1.0, 0.13, 0.32}
\definecolor{bostonuniversityred}{rgb}{1.0, 0.0, 0.0}

\usepackage[symbol]{footmisc}
\usepackage{hyperref}

\theoremstyle{definition}

\newcommand{\JS}[1]{{\color{black} #1}}
\newcommand{\HX}[1]{\textcolor{black}{#1}}

\journal{Elsevier}

\begin{document}

\begin{frontmatter}



\title{\textbf{Spatial-Temporal Selective State Space (ST-Mamba) Model for Traffic Flow Prediction}}


\author[first]{Zhiqi Shao}
\ead{zhiqi.shao@sydney.edu.au}
\author[first]{Michael G.H. Bell}
\ead{michael.bell@sydney.edu.au}
\author[first]{Ze Wang}
\ead{ze.wang@sydney.edu.au}
\author[first]{D. Glenn Geers}
\ead{glenn.geers@sydney.edu.au}
\author[third]{Haoning Xi\corref{cor1}}
\cortext[cor1]{Corresponding author}
\ead{alice.xi@newcastle.edu.au}
\author[first]{Junbin Gao}
\ead{junbin.gao@sydney.edu.au}

\affiliation[first]{organization={The University of Sydney Business School},
            addressline={Camperdown}, 
            city={Sydney},
            postcode={2006}, 
            state={NSW},
            country={Australia}}

\affiliation[third]{organization={Newcastle Business School, The University of Newcastle},
            addressline={Callaghan}, 
            city={Newcastle},
            postcode={2308}, 
            state={NSW},
            country={Australia}}

\begin{abstract}

Traffic flow prediction, a critical aspect of intelligent transportation systems, has been increasingly popular in the field of artificial intelligence, driven by the availability of extensive traffic data. The current challenges of traffic flow prediction lie in integrating diverse factors while balancing the trade-off between computational complexity and the precision necessary for effective long-range and large-scale predictions. To address these challenges, we introduce a Spatial-Temporal Selective State Space (\textsf{ST-Mamba}) model, which is the first to leverage the power of spatial-temporal learning in traffic flow prediction without using graph modeling. The \textsf{ST-Mamba} model can effectively capture the long-range dependency for traffic flow data, thereby avoiding the issue of over-smoothing. The proposed \textsf{ST-Mamba} model incorporates an effective Spatial-Temporal Mixer (ST-Mixer) to seamlessly integrate spatial and temporal data processing into a unified framework and employs a Spatial-Temporal Selective State Space (ST-SSM) block to improve computational efficiency. The proposed \textsf{ST-Mamba} model, specifically designed for spatial-temporal data, simplifies processing procedure and enhances generalization capabilities, thereby significantly improving the accuracy of long-range traffic flow prediction. Compared to the previous state-of-the-art (SOTA) model, the proposed \textsf{ST-Mamba} model achieves a 61.11\% improvement in computational speed and increases prediction accuracy by 0.67\%. Extensive experiments with real-world traffic datasets demonstrate that the \textsf{ST-Mamba} model sets a new benchmark in traffic flow prediction, achieving SOTA performance in computational efficiency for both long- and short-range predictions and significantly improving the overall efficiency and effectiveness of traffic management.




\end{abstract}




\end{frontmatter}




\section{Introduction}\label{Introduction}
Accurate and efficient traffic flow prediction is crucial for the 
enhancing road safety, reducing congestion, and improving the overall efficiency of intelligent transportation systems~\citep{Souza2017}. 
Although existing research has shown promise in short-range predictions~\citep{chen2011short, ProSTformer, CHEN2022522}, several challenges remain unresolved, such as the separate processing of spatial and temporal data~\citep{shao2024ccdsreformer, pdformer, huang2023adaptive, staeformer} and the difficulty in achieving long-range predictions while maintaining manageable computational complexity~\citep{CHEN2024102146, staeformer, ALHUTHAIFI2024120482}. Moreover, the previous models aiming to improve the precision of traffic flow predictions often resulted in higher computational complexities, compromising the efficiency of the prediction process. While faster algorithms can process large datasets more efficiently and yield more accurate predictions, the trade-off between speed and precision of prediction presents a significant challenge. High-accuracy algorithms typically require complex computations that can decelerate the prediction process, adversely impacting real-time decision-making. Therefore, it is necessary to develop robust traffic flow prediction models that ensure both high accuracy and computational efficiency, addressing these challenges effectively.

\HX{The emerging state space deep learning architecture, known as Mamba~\citep{gu2023mamba}, is increasingly recognized for its potential to surpass foundational models such as Graph Neural Networks (GNNs) \citep{GNNs2009}, and Transformers\citep{Vaswani2017attention}, in managing long-range, complex dynamic systems, thereby offering a promising solution to the persistent challenges in traffic prediction. In this paper, we introduce the first Spatial-Temporal Selective State Space (\textsf{ST-Mamba}) Model for traffic flow prediction, incorporating the innovative Spatial-Temporal Selective State Space block (ST-SSM block) with an ST-Mamba layer. The proposed \textsf{ST-Mamba} model efficiently integrates spatial and temporal data, balancing prediction accuracy and computational cost, particularly for long-range data. The core components of our model, the ST-Mixer and ST-Mamba layer, effectively combine spatial and temporal information, enhancing accuracy while reducing computational complexity. Extensive experiments with real-world traffic datasets demonstrate that the proposed \textsf{ST-Mamba} model excels at capturing long-range temporal dependencies and integrating spatial-temporal features. It outperforms existing benchmarks, achieving state-of-the-art performance on computational efficiency for both long-range and short-range forecasts. These results position the proposed \textsf{ST-Mamba} model as a breakthrough in traffic flow prediction, supporting more accurate and real-time decision-making in traffic control systems.}

We next review the literature on traditional models for traffic flow prediction (Section~\ref{tm}), deep learning methods for traffic flow prediction (Section~\ref{dlm}), and State Space Model (Section~\ref{ssm_intro}),
before outlining our contributions (Section~\ref{contr}).

\subsection{Traditional Models for Traffic Flow Prediction}\label{tm}
Traditional models for traffic flow prediction can be broadly classified into parametric and non-parametric methods. Parametric approaches like ARIMA models~\citep{ahmed1979analysis,zeng2008short,chen2011short,tong2008highway}, Kalman filters~\citep{KUMAR2017582,EMAMI2020102025,xu2017real,zhou2019hybrid,gao2013application}, artificial neural networks~\citep{topuz2010hourly,zeng2008short,sharma2018ann,cohen2019pedestrian}, regression techniques~\citep{alam2019prediction,priambodo2017predicting} focus heavily on temporal features but often overlook crucial spatial information vital for accurately predicting traffic flows in interconnected networks~\citep{EMAMI2020102025}. On the other hand, non-parametric machine learning techniques such as K-nearest neighbors~\citep{luo2019spatial-temporal,rahman2020short,cai2020sample,yang2019k}, and support vector machines~\citep{rahman2020short} have also been applied in traffic flow prediction but often struggle with the high dimensionality of traffic data, leading to inconsistent performance across diverse traffic conditions.

\subsection{Deep Learning Methods for Traffic Flow Prediction}\label{dlm}
In recent years, deep learning models like Recurrent Neural Networks (RNNs)\citep{Rumelhart1986rnn}, Convolutional Neural Networks (CNNs) \citep{CNNs1998}, Graph Neural Networks (GNNs)\citep{GNNs2009}, and Transformers\citep{Vaswani2017attention} have advanced traffic flow prediction.
However, each of these models has challenges that limit their effectiveness and, consequently, the overall efficiency and safety of transportation systems.

RNNs \citep{Rumelhart1986rnn}, especially Long Short-Term Memory (LSTM) networks \citep{hochreiter1997long}, can capture both long- and short-term dependencies but are susceptible to the vanishing gradient problem~\citep{hochreiterVanishingGradientProblem1998}. They are also computationally intensive~\citep{miglani2019deep, oliveira2021forecasting,luo2019spatial-temporal,ma2021short,LI2021102977,YANG2021103228}. Moreover, the sequential nature of RNNs \citep{Rumelhart1986rnn} restricts parallelization, leading to slower training and inference times, which limits their use in real-time traffic systems where rapid decision-making is crucial.

Similarly, CNNs \citep{CNNs1998} are highly effective at capturing spatial patterns in grid-like data and have been preferred for early-stage traffic prediction tasks~\citep{zhang2017stresnet, KE2017591, DUAN2016168}. However, they also face significant challenges in handling long-range temporal dependencies. While hybrid models combining CNNs \citep{CNNs1998} with other architectures like gated recurrent units (GRUs) have shown promise, they still require high inference time and computational resources, which makes them less practical for large-scale systems~\citep{WU2018166}. Furthermore, \cite{zhang2019short} demonstrated that while spatial-temporal feature selection algorithms enhance predictive performance, they still face challenges with high inference times and managing long-range temporal dependencies.

GNNs~\citep{GNNs2009} have gained popularity in traffic prediction due to the inherently interconnected nature of transportation networks, which can be effectively modeled as graphs. Models like GWNet~\citep{wu2020connecting} utilize GNNs to explore hidden spatial dependencies by automatically identifying directed relationships among variables. Similarly, Diffusion Convolutional Recurrent Neural Network (DCRNN)~\citep{li2017diffusion} models traffic flow as a diffusion process, combining convolutional and recurrent networks to capture spatial and temporal dynamics effectively. AGCRN~\citep{bai2020adaptive} introduces adaptive modules to improve sensitivity to node-specific patterns, demonstrating a sophisticated understanding of traffic's spatial-temporal dynamics. Meanwhile, STGCN~\citep{ijcai2018p505} merges graph convolutions with gated temporal convolutions to efficiently handle spatial and temporal variations. Additionally, GMAN~\citep{Zheng_Fan_Wang_Qi_2020} integrates spatial and temporal attention within a graph framework, showcasing the potential of attention mechanisms in traffic flow prediction. More recently, TFM-GCAM~\citep{CHEN2024102146} incorporates a traffic flow matrix to enhance the spatial-temporal features and dynamic characteristics of nodes. Despite their strengths, GNNs face challenges related to computational demands and modeling long-range dependencies. The over-smoothing problem, which occurs as node features become indistinguishable with increasing layers, makes it difficult to model long-range dependencies accurately and thus limits their predictive scope.

Recent research has explored the use of Transformer models~\citep{Vaswani2017attention} in traffic flow prediction. Transformers capture dependencies through self-attention mechanisms, allowing them to model complex relationships without the limitations as RNNs~\citep{Rumelhart1986rnn}. The transformer architecture has been successful in various spatial and temporal applications, such as NLP~\citep{devlin2018bert, zhang2020multi}, image classification~\citep{dosovitskiy2021an, caron2021emerging}, and video representation~\citep{girdhar2019video}. In traffic flow prediction, transformers have been applied through task-specific attention modules or direct architecture utilization~\citep{xu2020spatial, zheng2020gman, cai2020traffic, li2021adaptive}. Recent works have focused on improving transformer-based models for traffic flow prediction, such as the Adaptive Spatial-Temporal Transformer Graph Network (ASTTGN)~\citep{huang2023adaptive}, federated learning with adaptive spatial-temporal graph attention networks~\citep{ALHUTHAIFI2024120482}, and models capturing dynamic spatial dependencies and time delay~\citep{pdformer, shao2024ccdsreformer, staeformer}. However, Transformer models have their own challenges. The self-attention mechanism can be computationally expensive, especially for large-scale traffic networks and long sequences, with complexity growing quadratically with input length. This high computational cost can hinder practical application in long-term traffic management systems requiring swift prediction. Moreover, Transformers often need separate attention blocks for spatial and temporal information, limiting their ability to capture complex spatial-temporal dependencies. Developing a more integrated and efficient model remains an active research area.

\subsection{Selective State Space (Mamba) Model}\label{ssm_intro}

Inspired by the emerging Selective State Space model (so-called Mamba) introduced by~\citet{gu2023mamba}, which offers a viable solution that handles long-range time series data efficiently while maintaining high precision in prediction, we aim to address the existing challenges in traffic flow prediction by leveraging the power of Mamba model, with design an innovative ST-Mamba model enables the Mamba adoptive for spatial-temporal data. The efficiency of the Mamba model is valuable in both short-term and long-term traffic management, where fast and reliable predictions are critical for effective congestion control, route optimization, and traffic flow regulation.


State Space Models (SSMs) have emerged as promising alternatives in sequence modeling, particularly notable since the introduction of the Structured State Space Sequence Model (S4), first introduced by~\citet{ssms_gu}, represents a significant advancement in handling long sequence data. Central to its design is enhancing the HiPPO Matrix\citep{NEURIPS2020_102f0bb6}, which can notably improve long-term memory capabilities. S4 effectively combines the strengths of RNNs~\citep{Rumelhart1986rnn} and CNNs\citep{CNNs1998}, thus optimizing performance across a wide range of tasks. Building on the foundation of S4, the Mamba model further enhances adaptability through trainable parameters and introduces a parallel selective scan to minimize computational costs.  

Mamba has demonstrated success in natural language processing, computer vision, medical imaging, and graph learning~\citep{gu2023mamba, Wang2023, Zheng2023, Chen2023, Li2023}. Initially applied to language modeling and sequence-to-sequence tasks, Mamba showed impressive performance compared to transformers with linear complexity scaling~\citep{gu2023mamba}. Subsequent research explored Mamba's potential in vision tasks, such as image classification~\citep{Zhu2024}, object detection~\citep{Yang2024}, and medical image segmentation~\citep{Ma2024, Xing2024, Guo2024}, leveraging its efficiency in capturing long-range spatial dependencies.

\begin{table}[H]
    \centering
        \caption{Challenges in different models for traffic flow prediction.}
    \setlength{\tabcolsep}{2pt}
\renewcommand{\arraystretch}{1.2}
  \resizebox{\textwidth}{!}{%
    \begin{tabular}{ccccc}
        \toprule
        \multirow{2}{*}\textbf{Challenges} & \textbf{CNNs} & \textbf{RNNs} & \textbf{GNNs} & \textbf{Transformers} \\
        & \citep{CNNs1998} & \citep{Rumelhart1986rnn} & \citep{GNNs2009} & \citep{Vaswani2017attention}  \\
        \midrule
        Long-Range Data Handling & \checkmark & \checkmark & \checkmark & - \\
        \multirow{2}{*}{Computational Efficiency} & \checkmark & \checkmark & - & \checkmark \\
        & (Slow Inference) & (Limited Parallelization) & - & (High Computational Cost) \\
        Separate Spatial \& Temporal Processing & \checkmark & \checkmark & \checkmark & \checkmark \\
        \hline
    \end{tabular}
   } 
    \label{tab:model_issues}
\end{table}

\begin{table}[ht]
  \centering
    \caption{Comparison of different neural network architectures over training and testing phase.}
  \resizebox{\textwidth}{!}{%
  \begin{tabular}{@{}lcccl@{}}
    \toprule
    & Training Phase & Testing Phase & Additional Issue \\
    \midrule
    RNN (1986)~\citep{Rumelhart1986rnn} & Slow & Fast & Rapid Forgetting \\
    LSTM (1997)\citep{hochreiter1997long} & Slow & Fast & Forgetting \\
    Transformer (2017)~\citep{Vaswani2017attention}& Fast & Slow & Ram \& Time: O($n^2$) \\
    Mamba (2024)\citep{gu2023mamba} & Fast & Fast & Ram \& Time: O(n) \\
    \bottomrule
  \end{tabular}
 } 
  \label{intro_time}
\end{table}

\HX{Building on previous literature, we summarize the challenges faced by the current traffic flow prediction models in Table~\ref{tab:model_issues} and compare the computational efficiency of existing models in Table~\ref{intro_time}. Table~\ref{tab:model_issues} shows that existing models for traffic flow prediction struggle with issues such as the need to process spatial and temporal data separately~\citep{staeformer, pdformer, shao2024ccdsreformer}. Table~\ref{intro_time} demonstrates the difficulty of achieving high accuracy in long-range forecasts without incurring substantial computational costs~\citep{CHEN2024102146, staeformer, ALHUTHAIFI2024120482}.}

\subsection{Our Contributions}
\label{contr}
\HX{To address the challenges of integrating spatial and temporal data processing and improving prediction accuracy without compromising computational efficiency in traffic flow prediction, we propose an innovative Spatial-Temporal Selective State Space (\textsf{ST-Mamba}) model for traffic flow prediction, which can effectively handle high-dimensional data and capture long sequences without incurring high computational costs. The \textsf{ST-Mamba} model integrates a ST-Mixer with a ST-SSM block: the ST-Mixer unifies spatial and temporal data processing in a single step, while the ST-SSM block excels at identifying crucial traffic patterns over time, improving the long-range prediction accuracy. 
Additionally, the \textsf{ST-Mamba} model captures long-range temporal dependencies and balances comprehensive analysis with high precision, enabling real-time traffic management decisions and promoting safer and more efficient intelligent transportation systems.}

\HX{To the best of our knowledge, this paper represents the first attempt to apply The Selective State Space model in traffic flow prediction, setting a new benchmark. Our main contributions can be summarized as follows:}
\begin{itemize}
   
    \item \HX{The proposed \textsf{ST-Mamba} model seamlessly integrates spatial and temporal data processing into a unified framework, eliminating the need for separate data handling stages. This integration simplifies incorporation into transportation systems, enabling precise traffic flow predictions and proactive traffic management. The \textsf{ST-Mamba} model can support real-time traffic forecasting and ensure feasibility for implementation by transport regulators, even those with limited computational resources.}

    \item \HX{The proposed \textsf{ST-Mamba} model marks the first implementation of the ST-Mamba layer in traffic flow prediction, effectively increasing the computational efficiency in traffic flow prediction tasks. As a pioneer in assessing the performance of both the ST-Mamba layer and an Attention-based layer, our model demonstrates that a single ST-Mamba layer can perform comparably to three attention layers while significantly enhancing the processing speed. Without relying on GNNs to handle the spatial-temporal data, the proposed \textsf{ST-Mamba} model more effectively captures the long-range dependency for traffic flow data, thereby avoiding the issue of over-smoothing. }

    \item \HX{The proposed \textsf{ST-Mamba} model outperforms the previous state-of-the-art (SOTA) model, setting a new benchmark in traffic flow prediction for both long-range and short-range predictions. Extensive experimental results using real-world traffic flow datasets demonstrate that our model achieves a 61.11\% improvement in computational speed and a 0.67\% improvement in prediction accuracy. This improvement in traffic flow prediction is nontrivial in traffic management systems, where real-time response and effective decision-making are essential.}
    
\end{itemize}

\HX{The remainder of the paper is organized as follows: Section~\ref{Prelimanry} presents the problem statement, Section~\ref{Methodology} introduces the proposed \textsf{ST-Mamba} model, Section~\ref{Experiment} evaluates the performance of the \textsf{ST-Mamba} model on real-world datasets. Section~\ref{Discussion} discusses the implications of the model on traffic management across varying time frames. Finally, Section~\ref{Conclusion} summarizes our findings and suggests research directions.}

\section{Problem Statement}\label{Prelimanry}
\subsection{Notations}
\begin{figure}[ht]
    \centering
    \includegraphics[scale = 0.3]{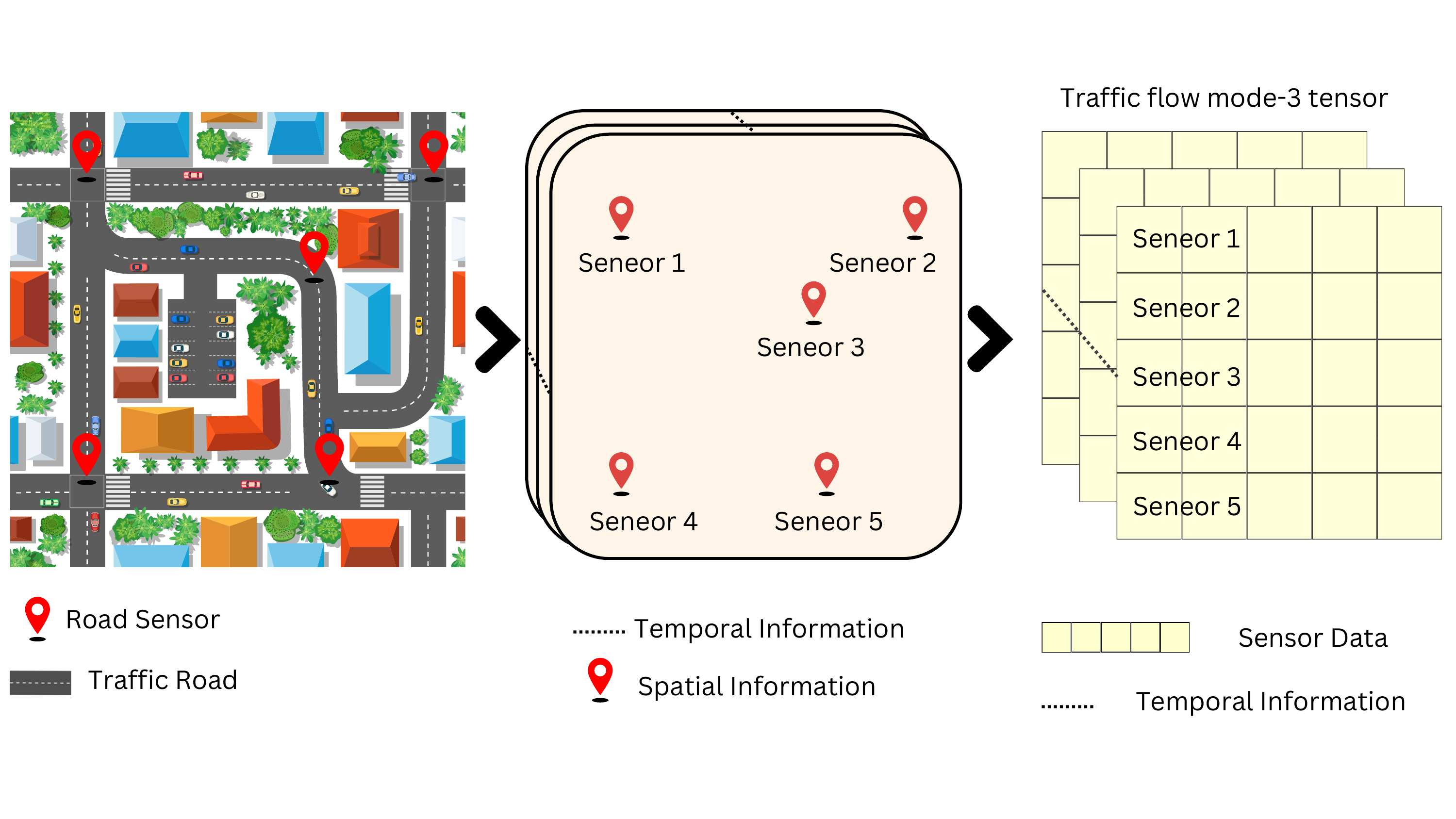}
    \caption{An illustration of the conversion of road maps into spatial-temporal information. }
    \label{fig:st-graph}
\end{figure}

Traffic flow prediction aims to forecast future traffic conditions accurately based on historical data. Given the traffic flow mode-3 tensor $\mathbf{X} \in \mathbb{R}^{T \times N \times d}$, structuring the temporal, spatial, and feature-specific dimensions of flow (Figure~\ref{fig:st-graph}), we train a model $f(\cdot)$ with parameters $\theta$ that utilizes the traffic flow information from the past $H$ timestamps to predict the traffic flow in the following $Z$ timestamps. This can be formally expressed as:
\begin{equation*}
[ \mathbf{X}_{t-H+1}, \ldots, \mathbf{X}_{t}]  \xrightarrow{f(\cdot; \theta)} [ \mathbf{X}_{t+1}, \ldots, \mathbf{X}_{t+Z}],
\end{equation*}
where $\mathbf{X}_{t}$ represents the traffic flow tensor at time $t$. We can set $t=H,\ldots, T-Z$ for the given traffic flow tensor $\mathbf{X}$ during the model learning stage. 

\HX{Figure~\ref{fig:st-graph} 
\JS{illustrates the process of collecting and structuring traffic data from road sensors for advanced traffic flow analysis. Road sensors, marked with red pins, are strategically placed at key intersections and road segments within a network to monitor traffic flow. These sensors gather data such as vehicle count and speed over time, capturing both spatial and temporal information. This collected data is then organized into a mode-3 tensor, where each layer represents data from a specific sensor (spatial dimension), and the entries within each layer correspond to different time points (temporal dimension). This tensor format allows for a comprehensive analysis of traffic patterns, enabling embedding this data into a structured format for advanced traffic analysis.}
}

We next introduce the foundational concepts and mathematical formulations that underpin the proposed traffic flow prediction model. Central to our approach is the adoption of The Selective State Space  model (Mamba) \citep{gu2023mamba}. 

\subsubsection{Selective State Space Models} \label{ssm}

\paragraph{State Equation} The Selective State Space model (Mamba) is a type of State Space models (SSMs) that offers a powerful framework for representing the dynamics of a system over time. These models fundamentally describe the evolution of a system's state and how observations or measurements are derived from these states. A discrete-time state space model is typically formulated using two principal equations: the state equation and the observation equation. The state equation describes how the state vector, \(\mathbf{H}_t\) from time step \(t-1\) to \(t\), which can be expressed as:
\begin{equation}\label{sse}
\mathbf{H}_t = \mathbf{A}_t \mathbf{H}_{t-1} + \mathbf{B}_t \mathbf{x}_t,
\end{equation}
where \(\mathbf{A}_t\) is the trainable state transition matrix with HiPPO matrix\citep{NEURIPS2020_102f0bb6} as initialization, and $\mathbf{B}_t$ is the trainable control-input matrix. The vector $\mathbf{x}_t$ is the control input at time step $t$. The state equation, as shown in Eq.~\eqref{sse}, captures the dynamics of the system by expressing the relationship between the current state, the control inputs, and the next state. It incorporates both internal and external factors that influence the system's state evolution. This comprehensive representation is pivotal for applications of Mamba in time series prediction, where the input $\mathbf x_t$ is derived from the observation at the previous timestamp $(t-1)$.

\paragraph{Observation Equation} On the other hand, the observation equation relates the output of observed traffic flow $\mathbf{y}_t$ at each time step \(t\) to the hidden state vector $\mathbf{H}_t$, given by:
\begin{equation}
\mathbf{y}_t = \mathbf{C}_t \mathbf{H}_t + \mathbf{v}_t,
\end{equation}
where $\mathbf{C}_t$ is the observation matrix, and $\mathbf{v}_t$ denotes the observation noise accounting for measurement errors or uncertainties. The observation equation defines how the underlying state of the system is reflected in the measurements or observations collected.

By combining the state and observation equations, Mamba enables the modeling of complex systems with hidden or unobserved states. Mamba is particularly suited for modeling traffic flow due to its ability to capture the temporal dependencies and dynamics of traffic flow, accommodate the stochastic nature of traffic data, handle incomplete or noisy observations, and provide a probabilistic framework for inferring the hidden states and making forecasts.

\paragraph{Selective Mechanism} The Selective State Space model (Mamba) ~\citep{gu2023mamba} incorporates an input-dependent mechanism that adapts and adjusts model parameters based on current inputs. It also features a selective mechanism designed to accelerate computational time. This enables the model to selectively focus on relevant information while disregarding extraneous details, thereby enhancing its capability to capture complex patterns in traffic data.

In the Mamba, the state transition matrix \(\mathbf{A}_t\), control-input matrix \(\mathbf{B}_t\), and observation matrix \(\mathbf{C}_t\) become functions of the input \(\mathbf{x}_t\):
\begin{equation}
\mathbf{A}_t = f_A(\mathbf{x}_t), \quad \mathbf{B}_t = f_B(\mathbf{x}_t), \quad \mathbf{C}_t = f_C(\mathbf{x}_t),
\end{equation}
where \(f_A\), \(f_B\), and \(f_C\) are learnable functions that map the input \(\mathbf{x}_t\) to the corresponding matrices. These functions can be implemented using neural networks, allowing the model to learn complex input-dependent transformations.

\section{Introduction of the \textsf{ST-Mamba} model}
\label{Methodology}
In this section, we provide an overview of the proposed \textsf{ST-Mamba} model (Section~\ref{ooverview}), introduce adaptive data embeddings (Section~\ref{Embedding}), ST-SSM block comprising the ST-Mixer, RMSNorm, Mamba layer and and regression layer (Section~\ref{ST-Mamba}).
\subsection{Overview}\label{ooverview}
\begin{figure}
    \centering
    \includegraphics[scale =0.3]{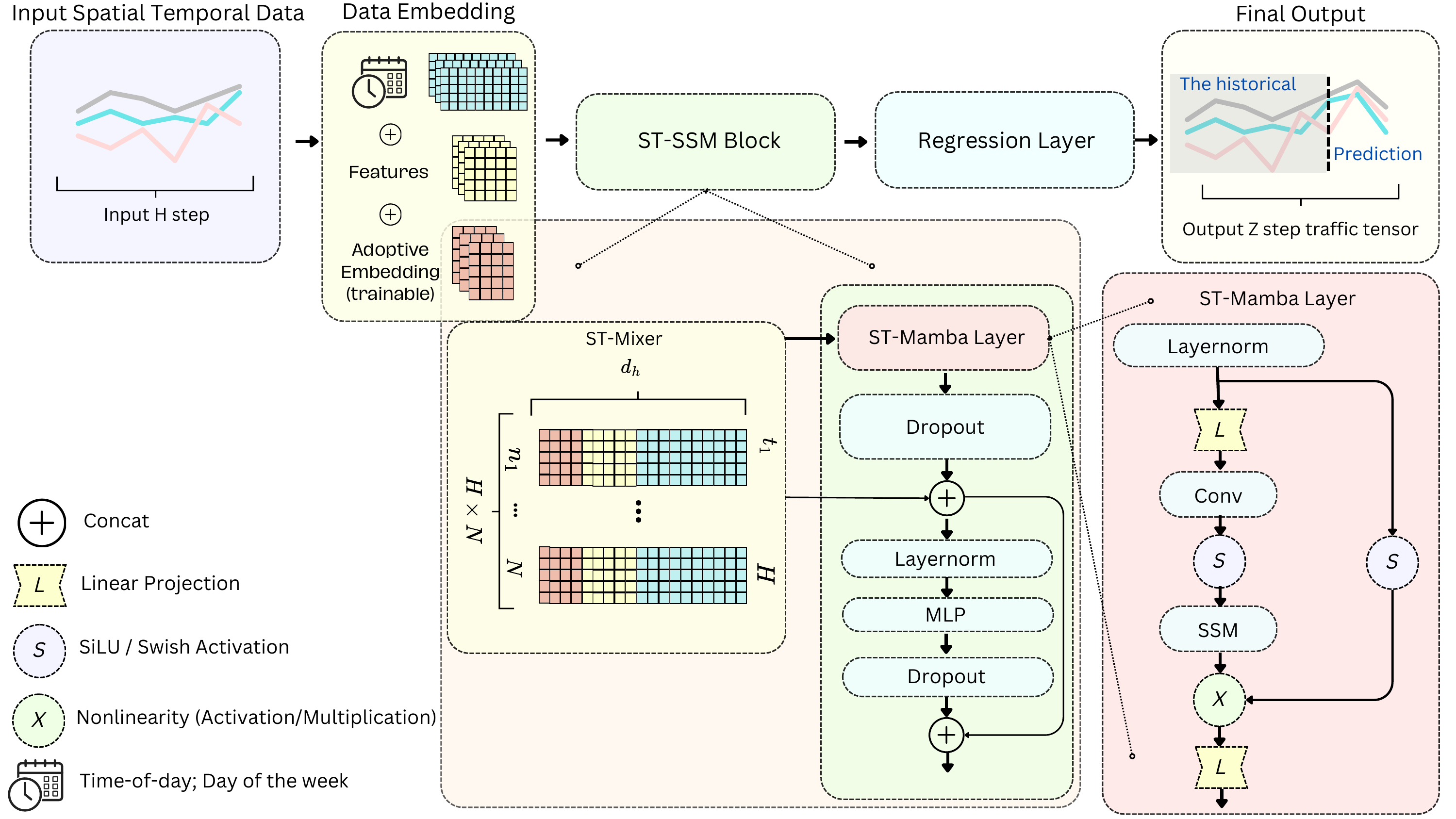}
    \caption{The framework of \textsf{ST-Mamba} model.}
    \label{fig:framework}
\end{figure}

The framework of \textsf{ST-Mamba} model is shown in Figure~\ref{fig:framework}. The input traffic flow data initially passes through the adaptive data embedding layer, which learns meaningful representations of the spatial and temporal features. 
Subsequently, the embedded data is reshaped and processed by the ST-Mixer, which seamlessly integrates spatial and temporal information. The integrated data is then fed into the ST-SSM block, a key component of the \textsf{ST-Mamba} architecture. The ST-SSM block, leveraging the power of The Selective State Space  Model (Mamba), efficiently captures the most significant patterns and long-range dependencies in the traffic flow data. Finally, the output of the ST-SSM block is passed through a regression layer, serving as a decoder to generate the final traffic flow prediction. This comprehensive design enables \textsf{ST-Mamba} model to deliver precise and reliable forecasts of traffic flow dynamics, effectively capturing the complex interplay between spatial and temporal factors in traffic flow.
 
\subsection{Data Embedding} \label{Embedding}
We employ the adaptive data embedding layer with input data $\mathbf{X}_{t{-H}+1:t}$. To retain the intrinsic information of the original data, we apply a fully connected layer to obtain the feature embedding $\mathbf{\tilde X}^f_t \in \mathbb{R}^{{H} \times N \times d_f}$:
\begin{equation}
\mathbf{\tilde X}^f_t = \text{FC}(\mathbf{X}
_{t{-H}+1:t})
\end{equation}
where $d_f$ represents the dimensionality of the feature embedding, and $\text{FC}(\cdot)$ denotes a fully connected layer~\citep{mlp2009}. We apply the learnable day-of-week embedding dictionary $\mathbf{\tilde X}^w_t \in \mathbb{R}^{N_w \times d_f}$ and the time-of-day embedding dictionary $\mathbf{\tilde X}^d_t \in \mathbb{R}^{N_d \times d_f}$, where $N_w = 7$ corresponds to the seven days of the week, and $N_d = 288$ represents the 288 timestamps (per 5 minutes) in a day. Let $\mathbf{W}_t \in \mathbb{R}^{H}$ and $\mathbf{D}_t \in \mathbb{R}^{H}$ denote the day-of-week and time-of-day data for the traffic series from $t{-H}+1$ to $t$. We use these as indices to retrieve the corresponding day-of-week embedding $\mathbf{\tilde X}^{t_w}_t \in \mathbb{R}^{{H} \times d_f}$ and time-of-day embedding $\mathbf{\tilde X}^{t_d}_t \in \mathbb{R}^{{H} \times d_f}$. By concatenating and broadcasting these embeddings, we obtain the periodicity embedding $\mathbf{\tilde X}^p_t \in \mathbb{R}^{{H} \times N \times 2d_f}$ for the traffic series. 

Traffic flow data is characterized by time's cyclic nature and the sequential flow of events, with periods often closely resembling to their immediate neighbors. Consequently, we utilized a shared spatial-temporal adaptive embedding as~\citep{staeformer}, $\mathbf{\tilde X}^a_t \in \mathbb{R}^{{H} \times N \times d_a}$, initialized with the Xavier\footnote{Xavier initialization sets initial weights to prevent extreme gradients early in training.} uniform initialization and then treated as a parameter of the model.

By combining the embeddings, we obtain the hidden spatial-temporal representation $\mathbf{\hat X_t} \in \mathbb{R}^{{H} \times N \times d_h}$:
\begin{equation}
\mathbf{\hat X_t} = \textbf{Concat}(\mathbf{\tilde X}^f_t , || , \mathbf{\tilde X}^p_t , || , \mathbf{\tilde X}^a_t)
\end{equation}
where $||$ means concatenate, and the hidden dimension $d_h$ is equal to $3d_f + d_a$.

\subsection{Spatial Temporal Selective State of Spatial (ST-SSM block)}\label{ST-Mamba}
As depicted in Figure~\ref{fig:framework}, an adaptive data embedding followed by an innovative ST-SSM block. The ST-SSM block, which contains an ST-Mamba layer, is designed to reduce the computational costs of the\textsf{ST-Mamba} model\@. Before implementing the ST-SSM block, it was essential to effectively integrate spatial and temporal information, necessitating the use of a data mixer known as the ST-mixer.

\subsubsection{ST-mixer} 

To effectively integrate spatial and temporal factors, \textsf{ST-Mamba} model utilizes tensor reshaping, detailed in Figure~\ref{fig:framework}, to convert tensor \( \mathbf{\hat X_t} \) into matrix \( \mathbf{\bar X_t} \), which is called matricization~\citep{TDA2009}. In our case, this is done by matricization along mode 3, denoted by,
\begin{align}
    \mathbf{\bar X_t} = reshape(\mathbf{\hat X_t}).
\end{align}
Through this reshaping, we obtain a new embedding matrix \( \mathbf{\bar X_t} \) in \( \mathbb{R}^{{(H\times N)} \times d_h} \). This reshaping facilitates the unified processing of spatial and temporal information, thereby simplifying and enhancing the ability of the  \textsf{ST-Mamba} model to capture complex traffic flow patterns more effectively. This also improves the generalization capabilities of the model. For simplicity, denote $T_1 = H\times N$ in the following discussion. 

\subsubsection{Normalization Layer} 
At the core of our ST-SSM block, we utilize the Layernorm, a normalization technique that enhances the stability and efficiency of training processes. Formally, given an input matrix \(\mathbf{\bar X} \in \mathbb{R}^{T_1 \times d_h}\), where \(T_1\) represents  the sequence length 
after merging the sequences of all the nodes, and \(d_h\) denotes the dimensionality of the feature space, LayerNorm is applied as follows:
\begin{align}
    \text{LayerNorm}(\mathbf{\bar X_t}) = \gamma \odot \frac{\mathbf{\bar X_t} - \mu}{\sqrt{\sigma^2 + \epsilon}} + \beta.
\end{align}
 Herein, \(\mu\) and \(\sigma^2\) are the mean and variance computed across the feature dimension \(d_h\), yielding vectors of dimensions \(T_1 \times 1\). The parameters \(\gamma\) and \(\beta\), with dimensions \(1 \times d_h\), serve as learnable scales and shifts, respectively, facilitating the adjustment of the normalization effect. This operation ensures that the model can maintain the benefits of normalization while retaining the capacity to recover the original distribution of activations. The inclusion of a small constant \(\epsilon\) ensures numerical stability by preventing division by zero. Through this mechanism, Layer Normalization contributes to mitigating the internal covariate shift\footnote{Internal covariate shift refers to changes in the distribution of network layer inputs during deep neural network training, caused by updates to layer weights. This shift can complicate training by requiring continuous adaptation of layers to new data distributions, potentially causing instability and slower convergence.}, thus accelerating the training convergence and enhancing the overall performance of deep neural architectures.

\subsubsection{ST-Mamba Layer} 

The ST-Mamba layer is based on the discretization of continuous SSMs, and the detail of the calculation is given in~\ref{proof1}. We present the input of the hidden state in the ST-SSM block as $\mathcal{H}_t$ where $\mathcal H_t = \text{LayerNorm}(\mathbf{\bar X_t})$ into The Selective State Space model (Mamba) layer for further processing. Herein the output after a linear projection of $\mathcal H_t$ is represented as $\mathbf{U}_t \in \mathbb{R}^{T_1 \times d_h}$ denote the hidden representation of the implicit latent state at time $t$. In ST-Mamba layer for iteration step $k$, we denote the input as $\mathbf{U}_t = [\mathbf{u}_1, \mathbf{u}_2, \dots, \mathbf{u}_{T_1}]^\top $, where $\mathbf{u}_k \in \mathbb{R}^{d_h}$, where $k = {1, 2, \dots, T_1}$ . The objective is to compute the prediction output of traffic flow $\mathbf{Y}_t \in \mathbb{R}^{T_1  \times d_h}$. The matrix $\mathbf{Y}_t$ comprises a sequence of prediction outputs \(\mathbf{y}_k\) for each time step \(k\) from 1 to \(T_1\). Specifically, we define \(\mathbf{Y}_t\) as specific $\mathbf{Y}_t = [\mathbf{y}_1, \mathbf{y}_2, \dots, \mathbf{y}_{T_1}]^\top$ for $k \in T_1$, where \(\mathbf{y}_k \in \mathbb{R}^{d_h}\) represents the prediction output at the step \(k\).

We initialize the ST-Mamba parameters and specify the dimensionality. Throughout this paper, we denote state dimension $d_\text{state}$. The state transition matrix \(\mathbf{A} \in \mathbb{R}^{d_\text{state} \times d_\text{state}}\) is established using HiPPO initialization~\citep{NEURIPS2020_102f0bb6} on matrix $\mathbf A$ to capture long-range dependencies. The matrix \(\mathbf{B} \in \mathbb{R}^{d_\text{state} \times d_h}\) is initialized as \(\mathbf{B}_1 = s_B(\mathbf{u}_1)\), where \(s_B(\cdot)\) represents a learnable linear projection. Throughout the iterative process, for each step \(k\), the matrix is updated according to \(\mathbf{B}_k = s_B(\mathbf{y}_k)\). Similarly, the output projection matrix \(\mathbf{C} \in \mathbb{R}^{d_h \times d_\text{state}}\) is initially computed as \(\mathbf{C}_1 = s_C(\mathbf{u}_1)\), with \(s_C(\cdot)\) also being a learnable linear projection. For each iteration step \(k\), the output projection matrix is updated to \(\mathbf{C}_k = s_C(\mathbf{y}_k)\). For each iteration, the output projection matrix on each step $k$ is $\mathbf{C}_k = s_B(\mathbf{y}_k)$. Then, the step size parameter denoted as $\Delta \in \mathbb{R}^{\mathbb{R}^{d_\text{state} \times d_h}}$ with the initialization $\Delta_1 = \tau_\Delta(\text{Parameter} + s_\Delta(\mathbf{u}_1))$. With each subsequent output \(\mathbf{y}_k\), the step size parameter for each iteration step \(k\) is updated as \(\Delta_k = \tau_\Delta(\text{Parameter} + s_\Delta(\mathbf{y}_k))\) where $\tau_\Delta$ is the softplus function\footnote{The softplus function is a smooth, differentiable activation function used in neural networks, defined as \( \text{softplus}(x) = \log(1 + e^x) \). It approximates the Rectified Linear Unit (ReLU) function but provides a continuous gradient, making it useful for gradient-based optimization. The derivative, \( \frac{d}{dx}(\text{softplus}(x)) = \frac{1}{1 + e^{-x}} \), ensures non-zero gradients for all input values, aiding in the prevention of vanishing gradients during training.} and with each iteration of $k$, the $s_\Delta(\cdot)$ is a learnable linear projection on the updated output of $y_k$. The dimensionality of $\Delta_k $ depends on whether the step size is uniform across all dimensions or varies per dimension. Notably, the parameter $\Delta_k $ in SSMs serves a similar function to the gating mechanism in RNNs. It controls the balance between how much the model should focus on the current input versus retaining information from previous states.

Next, we discretize the continuous-time parameters in \textsf{ST-Mamba} model (see details in~\ref{proof1}):
\begin{align}\label{eq:AB}
\mathbf{\tilde A}_k &= \exp(\Delta_k \mathbf{A}) \\
\mathbf{\tilde B}_k &= \mathbf{A}^{-1} (\exp(\Delta_k  \mathbf{A}) - \mathbf{I}) \mathbf{B}_k \approx (\Delta \mathbf{\tilde A}_k)(\Delta \mathbf{\tilde A}_k)^{-1}\Delta_k  \mathbf{\tilde{B}_k} = \Delta_k  \mathbf{B}_k
\end{align}

In practice as indicated in~\citep{gu2023mamba}, \(\Delta_k  \mathbf{B}_k \) is the solution of the approximation on \( \mathbf{\tilde{B}}_k  \) using the first-order Taylor series, and $\exp(\cdot)$ denotes the matrix exponential, and $\mathbf{I}$ is the identity matrix of appropriate size. The discretization operations apply to each time step, resulting in $\mathbf{\tilde A}_k \in \mathbb R ^{d_\text{state}  \times d_\text{state} }$ and $\mathbf{\tilde B}_k \in \mathbb R ^{d_\text{state} \times d_h}$. 

\begin{figure}[t]
    \centering
    \includegraphics[scale = 0.25]{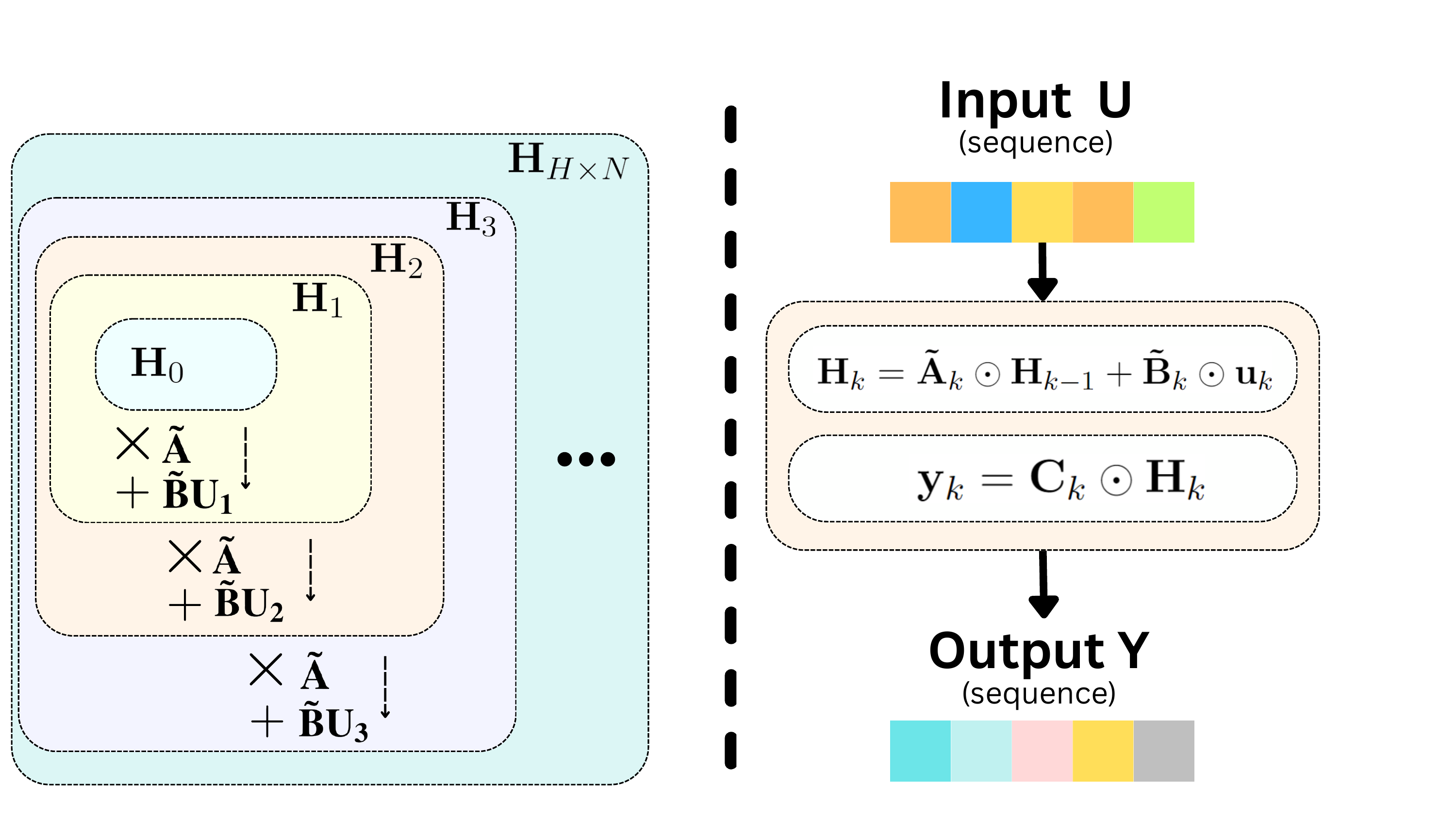}
    \caption{The intricacies of Mamba} 
    \label{fig:intrica_ssm}
\end{figure}

Then, we compute the traffic flow prediction output $\mathbf{Y}_t$ using the selective ST-Mamba layer recurrence. Figure~\ref{fig:intrica_ssm} displays the intricacies of Mamba: the left visual depicts the final hidden state, including the information from past states to enable long-term memory for long-range traffic flow data. The right visual offers a formulaic depiction of the Mamba, where each output $\mathbf Y_t$ incorporates historical information, which is suitable for long-term traffic flow prediction. We initialize the hidden state $\mathbf{H}_0 = \mathbf{0}$ with input $\mathcal{H}_t$ and iterate over each step $k = 1, \ldots, T_1$:
\begin{align}
\mathbf{H}_k &= \mathbf{\tilde A}_k \odot \mathbf{H}_{k-1} + \mathbf{\tilde B}_k \odot \mathbf{u}_k, \label{State_equ}\\
\mathbf{y}_k &= \mathbf{C}_k \odot \mathbf{H}_k , \label{State_equ2}
\end{align}
where $\odot$ represents the Hadamard product. Notably, for each iteration step $\mathbf{u}_k \in \mathbb{R}^{d_h}$, we illustrate it as the input of the ST-Mamba layer in Figure~\ref{fig:framework}. Here, $\mathbf{u}_k$ passes through a convolutional layer followed by a swish activation function. Subsequently, we apply the selective state output equation to obtain $\mathbf{y}_k \in \mathbb{R}^{d_{h}}$, computed by multiplying the hidden state $\mathbf{H}_k$ with the output projection matrix $\mathbf{C}_k$.
Each output $\mathbf{y}_k $ in step $k$ is then stacked and rearranged to $\mathbf{Y}_t \in \mathbb{R}^{T_1 \times d_{h}}$ as described in Eq.~\eqref{State_equ2}.

By leveraging the inherent advantages of the Mamba model~\citep{gu2023mamba}, the ST-Mamba layer has the capability efficiently to process long sequence data through a simplified hardware-aware parallel algorithm. Its selective state space mechanism effectively manages long sequences in the backward pass, which is essential for capturing long-term dependencies in traffic patterns. Unlike traditional Transformer models, the Mamba structure handles extended sequences with lower computational complexity, making it vital for tasks requiring long-term dependency modeling, such as traffic flow prediction.

\subsubsection{ST-SSM block and Regression layer} 
Once we obtain the predicted traffic flow output, denoted as \(\mathbf{Y}_t\), from the ST-Mamba layer, we will simplify notation by using \(\mathbf{Y}\) to represent \(\mathbf{Y}_t\) in subsequent discussions. Fine-tuning is then carried out using the ST-SSM block, which is
\begin{equation}
    \text{ST-Mamba}(\mathbf{Y}) = \text{Dropout}_2 \left( \text{MLP}\left( \text{LayerNorm}_1 \left( \text{Dropout}_1 (\mathbf{Y}) + \mathbf{\bar{X}} \right) \right) \right) + \left( \text{Dropout}_1 (\mathbf{Y}) + \mathbf{\bar{X}} \right).
\end{equation}
Here we perform a residual connection with $\left( \text{Dropout}_1 (\mathbf{Y}) + \mathbf{\bar{X}} \right)$ that helps in training deep networks by allowing gradients to propagate through the layers without diminishing. Also, applying layer normalization and dropout helps stabilize the learning process and prevents the model from overfitting to the training data. The whole ST-SSM block balances flexibility in fitting complex patterns in the data and stability in training. 
Finally, We utilized the regression layer as a decoder for the output of ST-SSM block which is in the space $\mathbb R^{T_1 \times d_h}$, here, we use an ST-Separator on the mixed spatial and temporal information with $\mathbf{\bar Y} =  reshape(\text{ST-Mamba}(\mathbf{Y}))$, where $\mathbf{\bar Y} \in \mathbb R^{{H} \times N \times d_h}$. The final transformation can be expressed as:
\begin{equation}
    \mathcal{Y} = \text{FC}(\mathbf{\bar Y}),
\end{equation}
where \(\mathcal{Y}\) is the final output, residing in the space \(\mathbb R^{Z \times N \times d}\), and $\text{FC}(\cdot)$ is the fully connected layer. This refined approach highlights architectural innovations to enhance deep network training and accurately decode complex, multidimensional data.


The Pseudocodes of the proposed ST-SSM block is given in Algorithm \ref{alg1}.
\begin{algorithm}[ht]
\caption{Pseudocodes of the proposed ST-SSM block.\label{alg1}}
\begin{algorithmic}[1]
\State \textbf{Input:} $\mathbf{U} \in \mathbb{R}^{\text{batch} \times T_1 \times d_h}$, the historical ground-truth ST data, where batch is batch size of model training, $T_1$ is the $H \times N$, $d_h$ is the total embedded feature dimension of data $\mathbf X$; 
\State \textbf{Output:} $\mathbf{Y}_t \in \mathbb{R}^{\text{batch} \times T_1\times d_h}$
\State \textbf{Params:} $A$ is the network param: nn.Parameter;
\State \textbf{Operator:} $s_B(\cdot)$, $s_C(\cdot)$ and $s_{\Delta}(\cdot)$ are the linear projection layer;
\State Initialize $\Delta, \mathbf B, \mathbf C$ using $\text{Linear}(\mathbf{u}_1)$.
\State $\mathbf{\tilde A}_k = \exp(\Delta A)$
\State $\mathbf{\tilde B}_k = (\Delta \mathbf A)^{-1}(\exp(\Delta \mathbf A) - I) \cdot \Delta \mathbf B$
\For{$k = 1$ to $T_1$}
    \State $h_k = \mathbf{\tilde A}_k  h_{k-1} + \mathbf{\tilde B}_k \mathbf u_k$ \Comment{ $\mathbf u_k$ is the input at step $k$} 
    \State $\mathbf y_k = \mathbf C_k h_k$ \Comment{$h_k$ is the hidden state at step $k$}
    \State \(\Delta_k = s_{\Delta}(\mathbf{y}_k)\)  \Comment{Update \(\Delta\) similarly}
    \State \(\mathbf{B}_k = s_B(\mathbf{y}_k)\)  \Comment{Update \(\mathbf{B}\) similarly}
    \State \(\mathbf{C}_k = s_C(\mathbf{y}_k)\)  \Comment{Update \(\mathbf{C}\) similarly}

\EndFor
\State \textbf{return} \(\mathbf{Y}_t = \text{stack}(\{\mathbf{y}_1, \mathbf{y}_2, \ldots, \mathbf{y}_{T_1}\})\)
\end{algorithmic}
\end{algorithm}

\section{Experimental Study}
\label{Experiment}
In this section, we first introduce the dataset and baseline models in Section~\ref{datadescription}. Next, we describe the experiment setup in Section~\ref{ExperimentSetup}. We then present the model performance evaluation for various datasets in Section~\ref{PerformanceEvaluation} and conduct an ablation study in Section~\ref{sec:Ablation}. Finally, we analyze the computational complexity in Section~\ref{Computational}.

\subsection{Data Description and Baseline Models}\label{datadescription}
\subsubsection {Datasets}

We evaluate the performance of the\textsf{ST-Mamba} model on six major traffic flow prediction benchmarking datasets, i.e., METR-LA, PEMS-BAY, PEMS03, PEMS04, PEMS07, and PEMS08, which are widely used in the traffic prediction research community. Table~\ref{data} summarizes the information of each dataset, wherein each dataset provides a rich source of information for model evaluation, including various traffic flow characteristics such as speed (mph), volume (vehicles per 5 minutes), and occupancy. 

\begin{table}[ht]
\scriptsize
\centering
\caption{Summary of traffic flow datasets}
\begin{tabular}{lcccccl}
\toprule
Datasets & \#Sensors & \#Records & Volume Units & Time Range       & Locations               \\ 
\midrule
METR-LA   & 207       & 34,272           & Vehicles/5min & 03/2012 - 06/2012 & Los Angeles             \\ 
PEMS-BAY & 325       & 52,116           & Vehicles/5min & 01/2017 - 05/2017 & San Francisco Bay Area  \\ 
PEMS03   & 358       & 26,209           & Vehicles/5min & 05/2012 - 07/2012 & California District 3   \\ 
PEMS04   & 307       & 16,992           & Vehicles/5min & 01/2018 - 02/2018 & California District 4   \\ 
PEMS07   & 883       & 28,224           & Vehicles/5min & 05/2017 - 08/2017 & California District 7   \\ 
PEMS08   & 170       & 17,856           & Vehicles/5min & 07/2016 - 08/2016 & California District 8   \\ 
\bottomrule
\end{tabular}
\label{data}
\end{table}

METR-LA \citep{li2017diffusion} comprises traffic speed data collected from 207 loop detectors across the LA County road network, spanning March to June 2012. Traffic data are recorded in 5-minute intervals, with a total of 34,272 records. PEMS-BAY \citep{li2017diffusion} includes traffic speed data from 325 sensors in the Bay Area, covering six months from January 1 to May 31, 2017. Traffic data are collected every 5 minutes, with a total of 52,116 records.  Compiled by the Caltrans Performance Measurement Systems (PeMS), PeMS03 \citep{song2020spatial}  are initially gathered every 30 seconds and aggregated into 5-minute intervals, featuring metrics such as speed, volume, and occupancy from March to June 2012.  PeMS04 \citep{song2020spatial} includes data from 307 sensors in the San Francisco Bay Area, spanning from January to February 2018 and it condenses data from each sensor into 5-minute intervals. PeMS07 \citep{song2020spatial} includes comprehensive traffic data from 883 sensors, recorded from May to August 2017. Each sensor's data is compiled into 5-minute intervals and includes traffic flow, speed, and occupancy measurements. PeMS08 \citep{song2020spatial}collects traffic data, including flow, speed, and occupancy, from 170 sensors between July and August 2016.

\subsubsection{Baseline Models} 

The performance of the proposed \textsf{ST-Mamba} model in traffic flow is compared with a set of comprehensive baseline models as follows,

\begin{itemize}
    \item Historical Index (HI)~\citep{cui2021historical} serves as the conventional benchmark, reflecting standard industry practices.
\end{itemize}
In particular, we examine a series of spatial-temporal Graph Neural networks, including
 
\begin{itemize}
    \item GWNet~\citep{wu2020connecting} presents a graph neural network framework designed to autonomously extract uni-directional relationships among variables, overcoming the limitations of previous approaches in harnessing latent spatial dependencies in multivariate time series prediction.
    \item DCRNN~\citep{li2017diffusion} develops the DCRNN for predicting traffic flow, effectively capturing both spatial and temporal dependencies.
    \item AGCRN~\citep{bai2020adaptive} offers adaptive modules that capture node-specific patterns and discern interdependencies among traffic series, enabling detailed modeling of spatial and temporal dynamics within traffic data.
    \item STGCN~\citep{ijcai2018p505} introduces a deep learning framework that combines graph convolutions for extracting spatial features with gated temporal convolutions for temporal features, thus enhancing both spatial and temporal analysis.
    \item GTS~\citep{shang2021discrete} develops a prediction approach for multiple interrelated time series, learning a graph structure in conjunction with a GNN, which addresses shortcomings of earlier methods.
    \item MTGNN~\citep{Wu2020ConnectingTD} introduces a graph neural network framework akin to GWNet, focusing on autonomously extracting uni-directional relationships among variables to capture both spatial and temporal dependencies.
    \item GMAN~\citep{Zheng_Fan_Wang_Qi_2020} introduces a graph-based deep learning model that employs spatial and temporal attention mechanisms to track dynamic correlations among traffic sensors effectively.
\end{itemize}
 
Recognizing the potential of Transformer-based models in time series prediction, we specifically focus on:

\begin{itemize}
\item PDFormer~\citep{pdformer} introduces a model for predicting traffic flow that captures dynamic and long-range spatial dependencies and the propagation delay of traffic conditions.
\item STAEformer~\citep{staeformer} proposes a spatial-temporal adaptive embedding that enhances the capabilities of standard transformers for traffic flow prediction, particularly effective in short-term prediction tasks.
\end{itemize}

Additionally, we explore further innovations in model architecture:
\begin{itemize}
\item STNorm~\citep{dengstnorm} utilizes spatial and temporal normalization modules to refine high-frequency and local components in the raw data.
\item STID~\citep{shaozhao2022} develops a method to enhance sample distinguishability in both spatial and temporal dimensions by integrating spatial and temporal identity information into the input data.
\end{itemize}

Comparing with the above diverse range of baseline models enables a robust validation for the performance of the proposed \textsf{ST-Mamba} model, ensuring comprehensive assessment across various dimensions in traffic flow prediction.

\subsection{Experiment Setup}
\label{ExperimentSetup}
\subsubsection{Implementation}
All experiments are conducted on a machine with the GPU RTX 3090(24GB) CPU 15. The dataset splits for PEMS-BAY, PEMS03, PEMS04, PEMS07, and PEMS08 are configured as follows: PEMS-BAY is segmented into training, validation, and test sets at a ratio of 7:1:2, while PEMS03, PEMS04, PEMS07, and PEMS08 follow a distribution ratio of 6:2:2. We set the embedding dimension (\(d_f\)) to 24 and the dimension of attention (\(d_a\)) to 80. The architecture includes 1 layer for the ST-Mamba layer, and the $d_\text{state}$ dimension for \textsf{ST-Mamba} model is 64. We define both the input and forecast horizon at 1 hour, corresponding to \({H} = Z = 12\). We employ the Adam optimizer for optimization, initiating with a learning rate of 0.001 that gradually decreases, and a batch size set at 16. To enhance training efficiency, an early stopping criterion is implemented, halting the process if validation error fails to improve over 30 consecutive iterations.

\subsubsection{Performance Metrics}
To evaluate the performance of traffic flow prediction methods, three prevalent metrics are employed: the Mean Absolute Error (MAE), the Mean Absolute Percentage Error (MAPE), and the Root Mean Square Error (RMSE). These metrics offer a comprehensive view of model accuracy and error magnitude. They are defined as follows:

   MAE (Eq.\eqref{MAE})  quantifies the average magnitude of the errors in a set of predictions without considering their direction. 
  \begin{equation}
    \text{MAE} = \frac{1}{n} \sum_{i=1}^{n} |\hat{y}_i - y_i|, \label{MAE}
  \end{equation}

MAPE (Eq.\eqref{MAPE}) expresses the error as a percentage of the actual values, providing a normalization of errors that is useful for comparisons across datasets of varying scales. 
  \begin{equation}
    \text{MAPE} = \frac{1}{n} \sum_{i=1}^{n} \left| \frac{\hat{y}_i - y_i}{y_i} \right| \times 100, \label{MAPE}
  \end{equation}

RMSE (Eq.\eqref{RMSE}) measures the square root of the average squared differences between the predicted and actual values, offering a high penalty for large errors. This metric is defined as:
  \begin{equation}
    \text{RMSE} = \sqrt{\frac{1}{n} \sum_{i=1}^{n} (\hat{y}_i - y_i)^2}, \label{RMSE}
  \end{equation}
where \(y = \{y_1, y_2, \ldots, y_n\}\) represents the set of ground-truth values, while \(\hat{y} = \{\hat{y}_1, \hat{y}_2, \ldots, \hat{y}_{n}\}\) denotes the corresponding set of predicted values. Through the utilization of MAE, MAPE, and RMSE, a thorough evaluation of model performance in prediction traffic conditions can be achieved, highlighting not just the average errors but also providing insights into the distribution and proportionality of these errors relative to true values.

\subsection{Performance Evaluation}
\label{PerformanceEvaluation}
In this section, we evaluate the performance of the proposed \textsf{ST-Mamba} model based on the six real-world datasets. Table~\ref{tab:performance_comparison_whole} presents the model performance on datasets PEMS03, PEMS04, PEMS07, and PEMS08, wherein the best results are highlighted as \textbf{bold}. The proposed \textsf{ST-Mamba} model outperforms the other models in datasets PEMS04, PEMS07, and PEMS08, standing out as a simpler, faster, and more effective solution. Specifically, the proposed \textsf{ST-Mamba} model outperforms the second-best model (the previous SOTA) model, STAEformer, in terms of MAE and RMSE on the PEMS04 dataset, with an improvement of 0.16\% and 0.03\%, respectively. On the PEMS08 dataset, \textsf{ST-Mamba} achieves a 0.45\% reduction in MAE and a 0.22\% reduction in RMSE compared to STAEformer. On the PEMS07 dataset, the proposed \textsf{ST-Mamba} model achieves the lowest MAE and RMSE values among all the compared models, with an MAE of 19.07 and an RMSE of 32.40. This represents a 0.37\% reduction in MAE and a 0.61\% reduction in RMSE compared to the second-best model, STAEformer, which has an MAE of 19.14 and an RMSE of 32.60. However, regarding MAPE, \textsf{ST-Mamba}'s performance (8.02\%) is slightly behind STAEformer (8.01\%) by a marginal difference of 0.01\%.  Among models that incorporate graph modeling, only the Graph WaveNet (GWNet) achieves a marginally higher MAE performance over the \textsf{ST-Mamba} model by 1.32 in the PEMS03 dataset.

\begin{table}[t!]
\centering
\caption{Comparative performance analysis of models on PEMS datasets.}
\label{tab:performance_comparison_whole}
\setlength{\tabcolsep}{3pt}
\resizebox{\textwidth}{!}{%
\begin{tabular}{l S[table-format=2.2] S[table-format=2.2] c S[table-format=2.2] S[table-format=2.2] c S[table-format=2.2] S[table-format=2.2] c S[table-format=2.2] S[table-format=2.2] c}
\toprule
{Model} & \multicolumn{3}{c}{PEMS03} & \multicolumn{3}{c}{PEMS04} & \multicolumn{3}{c}{PEMS07} & \multicolumn{3}{c}{PEMS08} \\
\cmidrule(lr){2-4} \cmidrule(lr){5-7} \cmidrule(lr){8-10} \cmidrule(lr){11-13}
& {MAE} & {RMSE} & {MAPE} & {MAE} & {RMSE} & {MAPE} & {MAE} & {RMSE} & {MAPE} & {MAE} & {RMSE} & {MAPE} \\
\midrule
HI & 32.62 & 49.89 & 30.60\% & 42.35 & 61.66 & 29.92\% & 49.03 & 71.18 & 22.75\% & 36.66 & 50.45 & 21.63\% \\
GWNet & \textbf{14.59}& \textbf{25.24} & 15.52\% & 18.53 & 29.92 & 12.89\% & 20.47 & 33.47 & 8.61\% & 14.40 & 23.39 & 9.21\% \\
DCRNN & 15.54 & 27.18 & 15.62\% & 19.63 & 31.26 & 13.59\% & 21.16 & 34.14 & 9.02\% & 15.22 & 24.17 & 10.21\% \\
AGCRN & 15.24 & 26.65 & 15.89\% & 19.38 & 31.25 & 13.40\% & 20.57 & 34.40 & 8.74\% & 15.32 & 24.41 & 10.03\% \\
STGCN & 15.83 & 27.51 & 16.13\% & 19.57 & 31.38 & 13.44\% & 21.74 & 35.27 & 9.24\% & 16.08 & 25.39 & 10.60\% \\
GTS & 15.41 & 26.15 & 15.39\% & 20.96 & 32.95 & 14.66\% & 22.15 & 35.10 & 9.38\% & 16.49 & 26.08 & 10.54\% \\
MTGNN & 14.85 & 25.23 & 14.55\% & 19.17 & 31.70 & 13.37\% & 20.89 & 34.06 & 9.00\% & 15.18 & 24.24 & 10.20\% \\
STNorm & 15.32 & 25.93 & 14.37\% & 18.96 & 30.98 & 12.69\% & 20.50 & 34.66 & 8.75\% & 15.41 & 24.77 & 9.76\% \\
GMAN & 16.87 & 27.92 & 18.23\% & 19.14 & 31.60 & 13.19\% & 20.97 & 34.10 & 9.05\% & 15.31 & 24.92 & 10.13\% \\
PDFormer & 14.94 & 25.39 & 15.82\% & 18.36 & 30.03 & 12.00\% & 19.97 & 32.95 & 8.55\% & 13.58 & 23.41 & 9.05\% \\
STID & 15.33 & 27.40 & 16.40\% & 18.38 & 29.95 & 12.04\% & 19.61 & 32.79 & 8.30\% & 14.21 & 23.28 & 9.27\% \\
STAEformer& 15.35 & 27.55 & 15.18\% & 18.22 & 30.18 & 11.98\% & 19.14 & 32.60 & \textbf{8.01\%} & 13.46 & 23.25 & \textbf{8.88\%} \\
\hline
\textsf{ST-Mamba}  & 15.28 & 27.28& \textbf{15.12\%} & \textbf{18.19} &\textbf{30.17} & \textbf{11.88\%} &\textbf{19.07}& \textbf{32.40}& 8.02\% & \textbf{13.40} & \textbf{23.20} & 9.00\% \\
\bottomrule
\end{tabular}
} 
\end{table}

 Table~\ref{tab:horizon} displays the performance evaluation of models on the METR-LA and PEMS-BAY datasets, considering three different horizons: 15 minutes, 30 minutes, and 60 minutes. These intervals represent short-term (15-min) and long-term predictions (30-min and 60-min). The best results are highlighted as {\textbf{bold}}.

For the PEMS-BAY dataset, the proposed spatial-temporal State Space (\textsf{ST-Mamba}) model exhibits superior performance across most prediction horizons.
For short-term prediction (15 minutes), the Graph WaveNet (GWNet) surpasses other models. This is likely because GNN-based models excel in capturing immediate spatial dependencies through their focus on nearest neighbors, which is advantageous for short-term prediction. However, the GNNs model is limited in capturing more localized spatial relations, and increasing their layers may lead to over-smoothing issues, diminishing model performance. This can be seen from the longer horizon time of 30-min and 60-min in PEMS-BAY. For a 30-min horizon, \textsf{ST-Mamba} improves over STAEformer(SOTA) by 0.62\% in terms of MAE, 9.78\% in terms of RMSE, and 1.66\% in terms of MAPE. At the 60-min horizon, \textsf{ST-Mamba} ties with STAEformer in terms of MAE, while STAEformer improves over \textsf{ST-Mamba} by 0.91\% in terms of RMSE, and \textsf{ST-Mamba} improves over STAEformer by 0.23\% in terms of MAPE.

From the observation on METR-LA, \textsf{ST-Mamba} model is competitive with the SOTA (STAEformer). Specifically, at the 15-min horizon, \textsf{ST-Mamba} improves over STAEformer by 0.38\% in terms of MAE and 3.50\% in terms of MAPE, while STAEformer improves over \textsf{ST-Mamba} by 1.16\% in terms of RMSE. At the 30-min horizon, STAEformer outperformed \textsf{ST-Mamba} by 1.00\% in terms of MAE and 2.91\% in terms of RMSE, while \textsf{ST-Mamba} improved over STAEformer by 0.86\% in terms of MAPE. At the 60-min horizon, \textsf{ST-Mamba}improves 0.30\% in terms of MAE, 1.13\% in terms of RMSE, and 1.44\% in terms of MAPE. For long-term prediction (60-minute horizon), the transformer-based model, e.g. STAEformer, with its three attention layers, shows competitive performance against the single ST-Mamba layer in the \textsf{ST-Mamba} model. This comparison suggests that a single layer of the \textsf{ST-Mamba} model is adequate for effective long-term traffic flow prediction, presenting a more efficient alternative to using three layers of attention. This insight underscores the efficiency of the \textsf{ST-Mamba} model in balancing model complexity and prediction accuracy, particularly for long-term prediction scenarios.

\begin{table}[t]
\centering
\caption{Comparative performance analysis of models on METR-LA and PEMS-BAY.}
\setlength{\tabcolsep}{0.8pt} 
\resizebox{\textwidth}{!}{%
\begin{tabular}{llcccccccccccccc}
\hline
 Horizon & Metric & HI & GWNet & DCRNN & AGCRN & STGCN & GTS & MTGNN & STNorm & GMAN & PDFormer & STID & STAEformer & \textsf{ST-Mamba} \\
\hline
\multicolumn{14}{c}{METR-LA} \\
\hline
15 min & MAE & 6.80 & 2.69 & 2.67 & 2.85 & 2.75 & 2.75 & 2.69 & 2.81 & 2.80 & 2.83 & 2.82 & 2.65 & \textbf{2.64}\\
& RMSE & 14.21 & 5.15 & 5.16 & 5.53 & 5.29 & 5.27 & 5.16 & 5.57 & 5.55 & 5.45 & 5.53 & \textbf{5.11} & 5.17\\
& MAPE & 16.72 & 6.99 & 6.86 & 7.63 & 7.10 & 7.12 & 6.89 & 7.40 & 7.41 & 7.77 & 7.75 & 6.85 &\textbf{6.61}\\
30 min & MAE & 6.80 & 3.08 & 3.12 & 3.20 & 3.15 & 3.14 & 3.05 & 3.18 & 3.12 & 3.20 & 3.19 & \textbf{2.97} & 3.00 \\
& RMSE & 14.21 & 6.20 & 6.27 & 6.52 & 6.35 & 6.33 & 6.13 & 6.59 & 6.49 & 6.46 & 6.57 & \textbf{6.00}& 6.18\\
& MAPE & 16.72 & 8.47 & 8.42 & 9.00 & 8.62 & 8.62 & 8.16 & 8.47 & 8.73 & 9.19 & 9.39 & 8.13 & \textbf{8.06}\\
60 min & MAE & 6.80 & 3.51 & 3.54 & 3.59 & 3.60 & 3.59 & 3.47 & 3.57 & 3.44 & 3.62 & 3.55 & 3.34 & \textbf{3.33}\\
& RMSE & 14.20 & 7.28 & 7.47 & 7.45 & 7.43 & 7.44 & 7.21 & 7.51 & 7.35 & 7.47 & 7.55 & \textbf{7.02} & 7.10\\
& MAPE & 10.15 & 9.96 & 10.32 & 10.47 & 10.35 & 10.25 & 9.70 & 10.24 & 10.07 & 10.91 & 10.95 & 9.70 & \textbf{9.56}\\
\hline
\multicolumn{14}{c}{PEMS-BAY} \\
\hline
15 min & MAE & 3.06 & \textbf{1.30} & 1.31 & 1.35 & 1.36 & 1.37 & 1.33 & 1.33 & 1.35 & 1.32 & 1.31 & 1.31 & \textbf{1.30}\\

& RMSE & 7.05 & \textbf{2.73} & 2.76 & 2.88 & 2.88 & 2.92 & 2.80 & 2.82 & 2.90 & 2.83 & 2.79 & 2.78 & 2.89\\

& MAPE & 6.85 & \textbf{2.71} & 2.73 & 2.91 & 2.86 & 2.85 & 2.81 & 2.76 & 2.87 & 2.78 & 2.78 & 2.76 & 2.94\\

30 min & MAE & 3.06 & 1.63 & 1.65 & 1.67 & 1.70 & 1.72 & 1.66 & 1.65 & 1.65 & 1.64 & 1.64 & 1.62 & \textbf{1.61}\\

& RMSE & 7.04 & 3.73 & 3.75 & 3.82 & 3.84 & 3.86 & 3.77 & 3.77 & 3.82 & 3.79 & 3.73 & 3.68 & \textbf{3.32}\\

& MAPE & 6.84 & 3.73 & 3.71 & 3.81 & 3.79 & 3.88 & 3.75 & 3.66 & 3.74 & 3.71 & 3.73 & 3.62 & \textbf{3.56}\\

60 min & MAE & 3.05 & 1.99 & 1.97 & 1.94 & 2.02 & 2.06 & 1.95 & 1.92 & 1.91 & 1.91 & 1.91 & \textbf{1.88} & \textbf{1.88}\\

& RMSE & 7.03 & 4.60 & 4.60 & 4.50 & 4.63 & 4.60 & 4.50 & 4.45 & 4.49 & 4.43 & 4.42 & \textbf{4.34}& 4.38\\

& MAPE & 6.83 & 4.71 & 4.68 & 4.55 & 4.72 & 4.88 & 4.62 & 4.46 & 4.52 & 4.51 & 4.55 & 4.41 & \textbf{4.40}\\
\hline
\end{tabular}
} 
\label{tab:horizon}
\end{table}

\subsection{Ablation Study}
\label{sec:Ablation}
In this analysis, we evaluate the performance of the \textsf{ST-Mamba} model in comparison to the Transformer-based model, utilizing the state-of-the-art model STAEformer as a benchmark. Notably, while both the STAEFormer and \textsf{ST-Mamba} models employ similar architectures, they differ in their encoding blocks. The STAEFormer comprises two attention blocks for processing temporal and spatial information, each consisting of three layers. On the other hand, the \textsf{ST-Mamba} model incorporates a single ST-SSM block with a single ST-Mamba layer without segregating the information encoding process.

\HX{Table~\ref{tab:Ablation} presents a comparison of the performance metrics for two models, the Transformer-based model (STAEformer) and the proposed Mamba-based model (ST-Mamba), as the number of layers is varied from 1 to 3. The metrics evaluated are Mean Absolute Error (MAE), Root Mean Squared Error (RMSE), and Mean Absolute Percentage Error (MAPE).}
where the best results are highlighted as {\textbf{bold}}, reveals that the \textsf{ST-Mamba} model, furnished with a ST-Mamba layer, outperforms the STAEformer (the original model with three layers) across various layer configurations. Notably, while increasing the number of attention layers in STAEformer amplifies its Mean Absolute Error (MAE), with a three-layer configuration yielding an MAE of 13.49, the \textsf{ST-Mamba} model achieves an even lower MAE of 13.40 with a ST-Mamba layer.

\begin{table}[ht!]
\centering
\setlength{\tabcolsep}{6pt}
\caption{Comparison of performance on the PEMS08 with various layers in STAEformer and \textsf{ST-Mamba} models.}
\label{tab:performance_comparison}
\begin{tabular}{ccccccc}
\toprule
Models & MAE & RMSE & MAPE \\ \hline
STAEformer (3 layers) & 13.49 & 23.30 & 8.84  \\ 
STAEformer (2 layers) & 13.54 & 23.47 & 8.887  \\ 
STAEformer (1 layer) & 13.77 & 23.27 & 9.16 \\ \hline
 \textsf{ST-Mamba} (3 layers) & 13.45& \textbf{23.08} & \textbf{8.96}  \\
 \textsf{ST-Mamba} (2 layers) & 13.43 & 23.14 & 8.95\\
 \textsf{ST-Mamba} (1 layer) & \textbf{13.40} & 23.20 & 9.00 \\
    \bottomrule
\end{tabular}\label{tab:Ablation}
\end{table}
\begin{figure}[ht!]
    \centering
    \includegraphics[scale = 0.47]{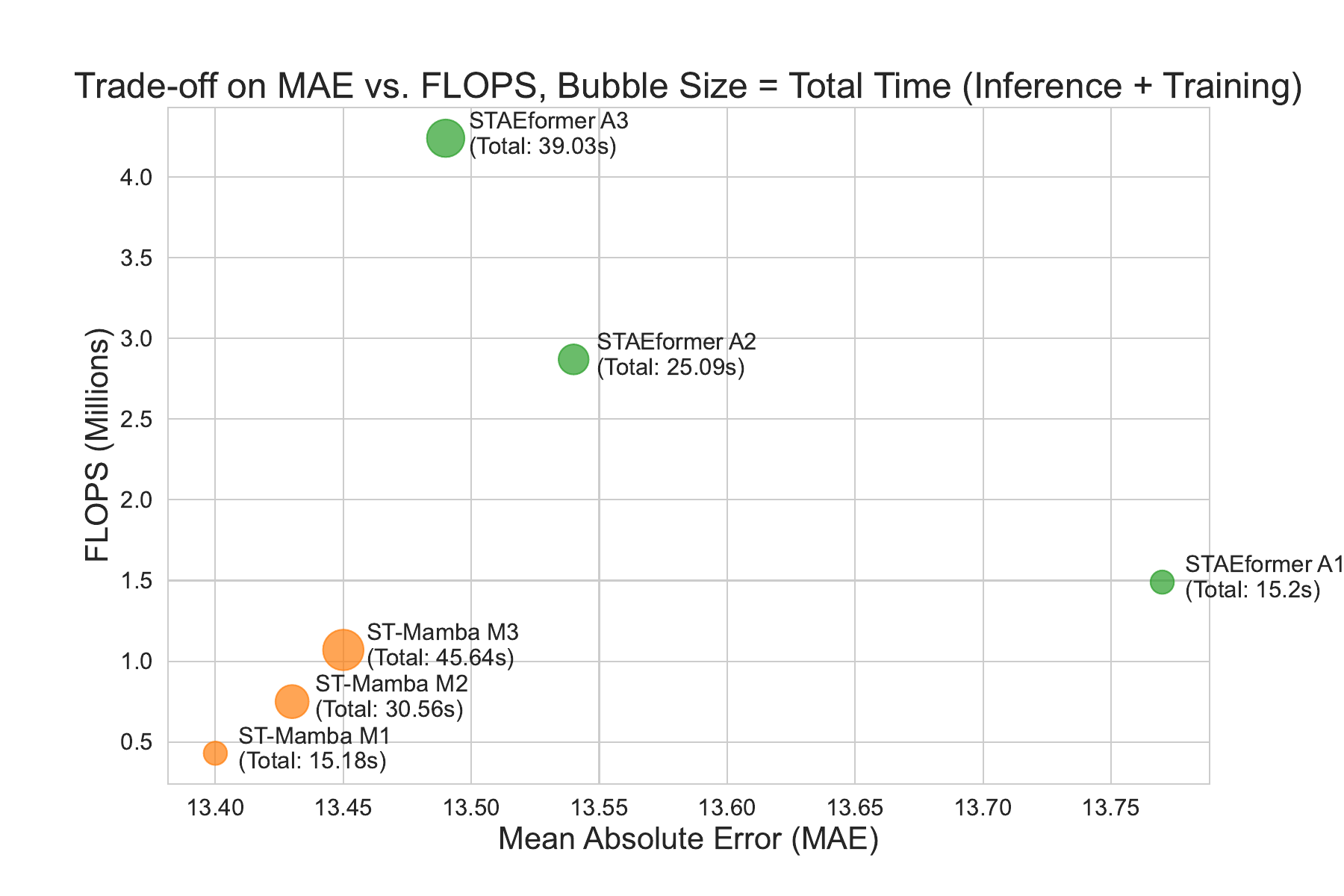}
    \caption{Comparison of FLOPS vs total computational time on ST-Mamba and STAEformer in PEMS08.}
    \label{fig:training_time}
\end{figure}

The trade-off on computational efficiency and accuracy illustrated in Figure~\ref{fig:training_time} which represents the results of PEMS08 with the comparison of computational metrics---FLOPS (Floating Point Operations Per Second, in millions), total computational time (inference time (in seconds) and training time (in seconds), among different transformer layers of the STAEformer model, and ST-Mamba layers in ST-Mamba\@. We denote ``M'' as the number of Mamba layers in the model,  ``A'' as the number of attention layers. The visualization highlights the impact of increasing the number of attention layers on the computational demands of training. A detailed comparison between the ST-Mamba, which incorporates a single ST-Mamba layer, and the STAEformer configured with one attention layer demonstrates that the Mamba-based approach significantly reduces training time while achieving competitive inference speeds, as shown in Figure~\ref{fig:training_time}. 

This analysis emphasizes the dual strengths of \textsf{ST-Mamba} model in improving prediction accuracy and computational efficiency, positioning it as a powerful tool for traffic flow prediction and related challenges in spatial-temporal data modeling. The \textsf{ST-Mamba} model stands out as an attractive solution for delivering timely and cost-effective predictions, excelling in both efficiency and effectiveness in processing and analyzing large datasets.


\begin{figure}[ht!]
    \centering
    \includegraphics[scale = 0.35]{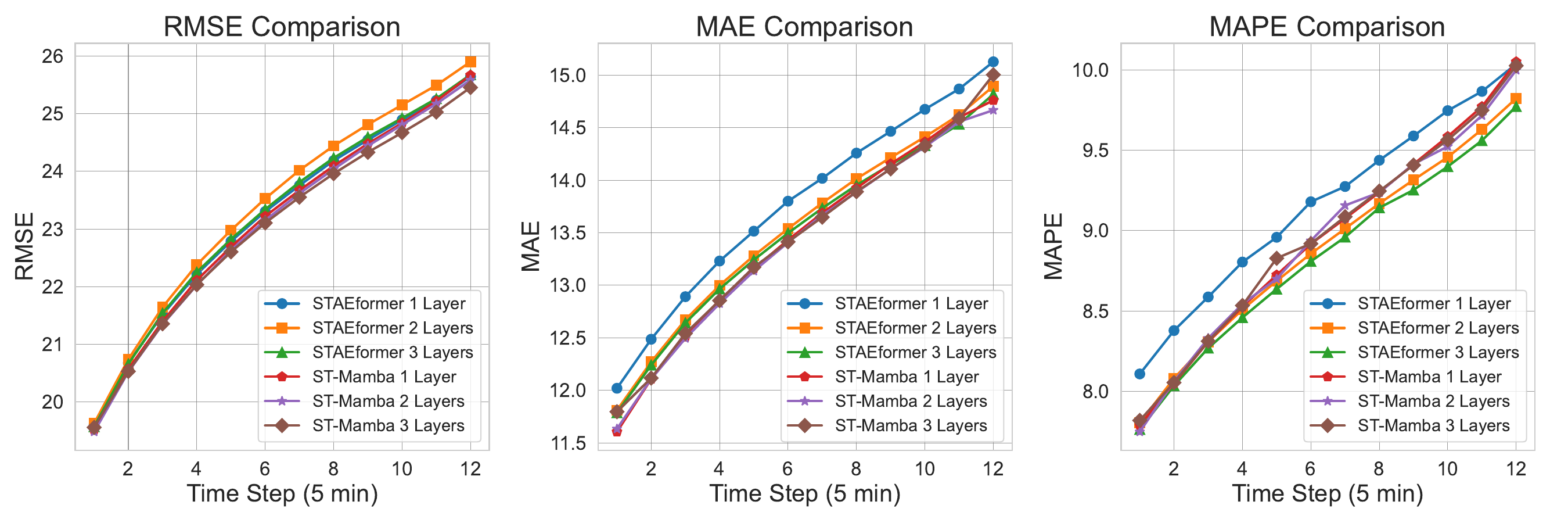}
    \caption{Comparison of RMSE, MAE, and MAPE across various layers for STAEformer and ST-Mamba.}
    \label{fig:predict_timestep}
\end{figure}

Next, we present a temporal analysis of traffic flow prediction with an hour-long forecast for PEMS08, which aims to delve into the temporal dynamics of traffic flow prediction by evaluating the model's performance at different time intervals. Figure~\ref{fig:predict_timestep} presents a side-by-side comparison of three key performance metrics, RMSE, MAE, and MAPE, across varying layers (1 Layer, 2 Layers, and 3 Layers) for two distinct models, STAEformer and ST-Mamba\@. Each subplot illustrates the variation of a specific metric across 12-time steps, highlighting the models' performance stability and accuracy in prediction. Distinct color-coded lines represent different model configurations, ensuring clear differentiation and readability. The visualization of prediction at each 5-minute time step for the metrics RMSE, MAE, and MAPE\@. The RMSE trends indicate a typical increase in prediction error as the interval extends. Notably, the \textsf{ST-Mamba} model exhibits a more gradual increase in RMSE across all layers, suggesting enhanced long-term prediction capabilities which are crucial for strategic traffic planning. The MAE results follow a similar upward trajectory for both models, signifying greater errors with longer forecasts. However, the ST-Mamba, particularly in the 1 and 2-layer configurations, shows a less steep increase except at the final time step. Interestingly, adding layers to the ST-Mamba does not significantly alter the outcome, hinting at its proficiency in retaining prediction accuracy over time. As for MAPE, the values rise with longer prediction intervals. The ST-Mamba layers maintain a steady progression of error, suggesting a consistent percentage error in relation to the actual traffic volume. Conversely, the 2 and 3-layer configurations of the STAEformer seem to adapt better to varying traffic conditions, which may offer advantages during times of fluctuating traffic volumes, like rush hours or quieter periods.

\subsection{Computational Complexity}
\label{Computational}

There are notable differences when comparing the computational complexity of the ST-Mamba with one layer of Mamba and the STAEformer model with two transformers, each having three layers. The \textsf{ST-Mamba} model's complexity scales linearly with the sequence length $T_1$, resulting in a complexity of $\mathcal{O}( B \times T_1 \times d_h \times (d_\text{conv} + d_\text{state}))$, $B$ is the batch size, $d_h$ is the hidden dimension size, $d_\text{state} = d_h \times expand$ and $expand$ is the expansion factor normally we use 2, $d_\text{conv}$ is the convolutional kernel size, and $d_\text{state}$ is the state dimension size. On the other hand, the STAEformer's complexity has a quadratic term $T_1^2$, resulting in a complexity of $\mathcal{O}(B\times T_1 \times d_h^2 + B \times T_1^2 \times d_h)$. This implies that as the sequence length $T_1$ increases, the complexity of \textsf{ST-Mamba} grows more slowly compared to the STAEformer. In practice, the ST-Mamba performs well with relatively small values for the additional factors, such as the expansion factor, convolutional kernel size, and state dimension size, compared to the hidden dimension size. Consequently, the ST-Mamba tends to have lower computational complexity for longer sequences than the STAEformer, making them more efficient in real-world scenarios.

\section{Discussion and Implication}
\label{Discussion}
\begin{figure}[ht!]
    \centering
    \includegraphics[scale = 0.31]{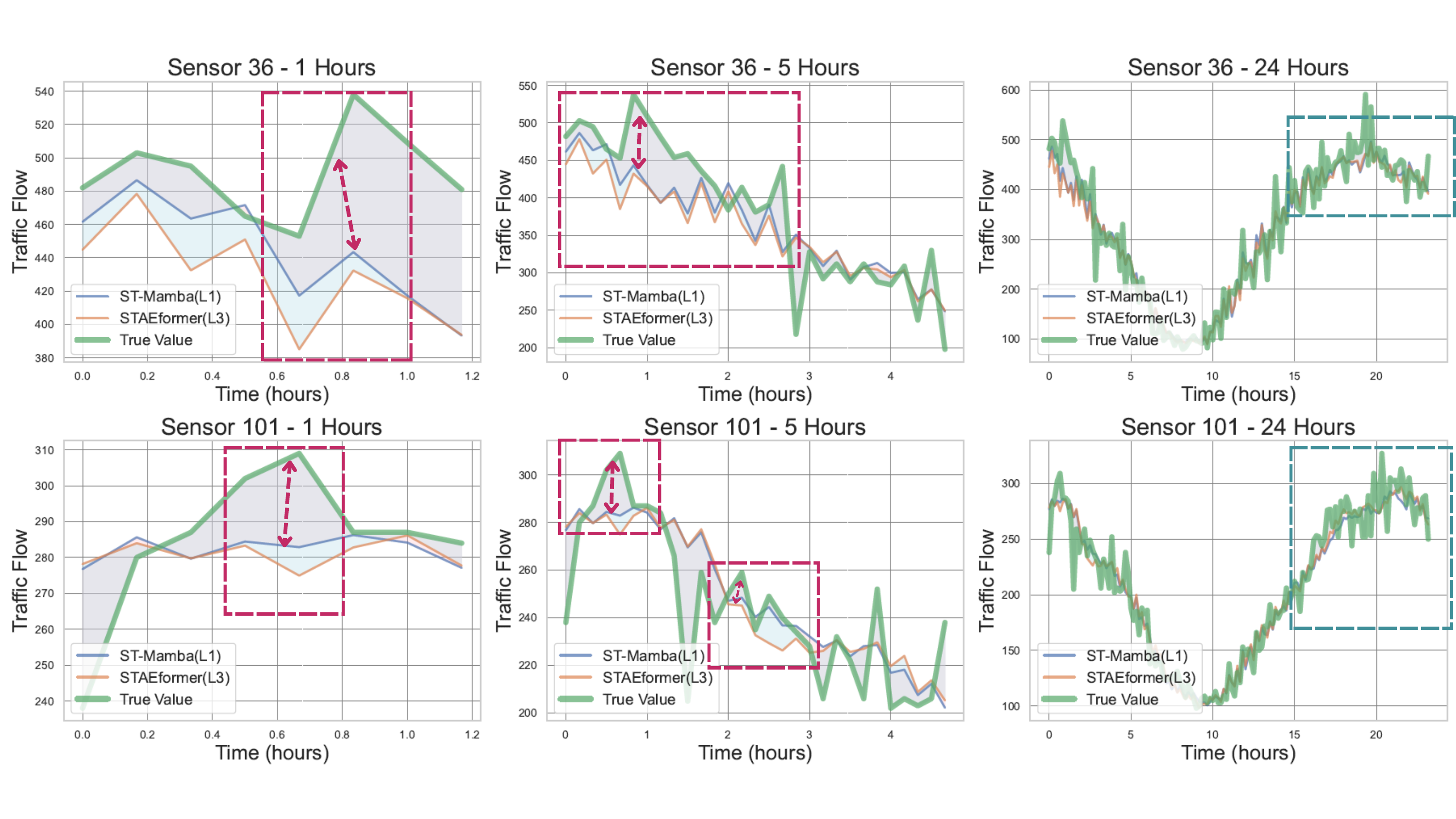}
    \caption{Comparison of STAEFormer and \textsf{ST-Mamba} in traffic flow prediction on PEMS08.}
    \label{fig:trueandpred}
\end{figure}

In this section, we present a comparative visualization of traffic flow predictions over time in Figure~\ref{fig:trueandpred}, focusing on measurements from two sensors—Sensor 36 and Sensor 101—at intervals of 1 hour, 5 hours, and 24 hours. The plots contrast the actual traffic flow values with predictions from two models: \textsf{ST-Mamba}  (L1) and STAEFormer(L3). Notably, \textsf{ST-Mamba}  (L1) incorporates a single Mamba layer, while STAEFormer(L3) is equipped with three attention layers. 

As shown in Table \ref{tab:horizon}, STAEFormer is the previous SOTA model (current second best model) in the baseline evaluation. In this section, in Figure~\ref{fig:trueandpred},  we present a comparative visualization of traffic flow predictions  measured by two sensors (Sensor 36 and Sensor 101) over different time intervals (1 hour, 5 hours, and 24 hours) in Figure~\ref{fig:trueandpred}. The plots compare the true valuesof traffic flow with two prediction models: \textsf{ST-Mamba} (L1) and STAEformer(L3) where \textsf{ST-Mamba}  (L1) uses one Mamba layer, and STAEFormer (L3) uses three attention layers.  

Regarding the 1-hour and 5-hour predictions, the \textsf{ST-Mamba} model significantly reduces trend deviations between predicted values and the actual data for both short and long-term forecasts. This underscores the efficacy of utilizing a single Mamba layer to manage diverse prediction scenarios. Notably, during periods of rapid fluctuations, marked by red dashed boxes, the \textsf{ST-Mamba} model demonstrates superior adaptability by aligning closely with the ground truth, thus showcasing its responsiveness to sudden changes.

The 24-hour predictions, indicated by green dashed boxes, suggest that both the \textsf{ST-Mamba} and STAEFormer models could benefit from enhancements, especially during periods of rapid traffic flow changes. The swift and frequent variations during these extended times pose a significant challenge. This observation points to a potential area for further model refinement to more accurately capture the complex dynamics of traffic patterns over longer durations.

To further explore the implications of the comparative analysis between the \textsf{ST-Mamba} and STAEFormer models for traffic management, it is essential to consider the following key managerial insights:

\begin{itemize}
\item The \textsf{ST-Mamba} model's proficiency in minimizing trend deviations for 1-hour and 5-hour forecasts translates into enhanced operational capabilities for dynamic traffic control measures. These include adaptive traffic signal control and real-time rerouting advice, which can respond more swiftly and accurately to changing traffic conditions. The model's ability to closely match actual traffic flow during rapid fluctuations further supports its utility in developing responsive traffic management strategies like enhanced incident detection and quicker response implementations.

\item While the \textsf{ST-Mamba} model shows potential for informing long-term traffic strategies through its 24-hour forecasts, both it and the STAEFormer model exhibit challenges in capturing the full complexity of traffic dynamics over such extended periods. This insight is crucial for strategic planning, including optimizing the timing and placement of road works and managing lane closures. It highlights the need for continuous model improvement to enhance accuracy in long-term traffic forecasting, which is essential for planning peak traffic flows and minimizing congestion and environmental impacts.

\item The superior adaptability of the \textsf{ST-Mamba} model during periods of rapid fluctuation underscores the value of further developing models that can effectively manage high-frequency variability in traffic patterns. By integrating more advanced deep learning techniques that excel in handling such variability, traffic management systems can become more proactive, not just in improving traffic safety but also in reducing the economic costs associated with congestion.
\end{itemize}

\section{Conclusion}
\label{Conclusion}
This paper proposes an innovative \textsf{ST-Mamba} model for traffic flow prediction, which can significantly improve computational efficiency by incorporating a spatial-temporal selective state space (ST-SSM) block. Extensive experimental results demonstrate that the proposed \textsf{ST-Mamba} model achieves state-of-the-art computational efficiency, with a \textbf{61.11\%} improvement in computational speed and a \textbf{0.67\%} increase in prediction accuracy, thereby facilitating more precise and real-time decision-making in transportation system management. Additionally, the proposed \textsf{ST-Mamba} model utilizes an ST-Mixer to integrate spatial and temporal data into a unified framework and simplifies the data processing procedure, making it more accessible for transport regulators, even those with limited computational resources. 
Extensive experiments show that the proposed \textsf{ST-Mamba} model can improve prediction accuracy and computational efficiency in long- and short-range traffic predictions. Specifically, compared with the STAEFormer \JS{(one of the SOTA models)}, the \textsf{ST-Mamba} model demonstrates an average improvement in MAE by \textbf{0.29\%}, RMSE by \textbf{1.14\%}, and MAPE by \textbf{0.61\%}. These achievements mark the proposed \textsf{ST-Mamba} model as a new benchmark in traffic prediction, providing a powerful tool for data-driven management in intelligent transportation systems.

Future research on the \textsf{ST-Mamba} model could explore several promising avenues further to improve its capabilities and applications in traffic flow prediction. Integrating additional data sources, such as weather conditions or event schedules, could refine the accuracy of the proposed \textsf{ST-Mamba} model under diverse environmental conditions, potentially leading to innovations in autonomous traffic management. Developing adaptive algorithms that dynamically respond to changes in traffic patterns could improve the real-time responsiveness of the \textsf{ST-Mamba} model. 
\section*{Acknowledgments}
This research is partially supported by UON CHSF 2024 Pilot Research Scheme (G2400027) and CHSF 2024 New Start Scheme (G2400081).

 \appendix
 \label{Appendix}
\section{Algorithim of discretization in SSM}\label{proof1}

Discretization is essential for transforming continuous models into discrete versions for computational analysis. It utilizes the exact solutions of a system's ODEs, ensuring that state space models (SSMs) accurately reflect the dynamics of the continuous process in discrete-time studies.

\subsection{Continuous-Time State Space Model}
Consider the continuous-time state space model with the following input:
\begin{align}
\dot{h} &= Ah + Bu, \label{eq:6.1} \\
y &= Ch, \label{eq:6.2}
\end{align}
where \( \dot{h} \) denotes the time derivative of the state vector \( h \), and equation \eqref{eq:6.1} is termed the state equation while equation \eqref{eq:6.2} is referred to as the measurement equation. Here, \( A \), \( B \), and \( C \) are matrices that define the system dynamics, input influence, and output relation, respectively.

The general solution to the state equation, which is a linear ordinary differential equation (ODE), is given by:
\begin{equation}\label{eq:6.3}
h(t) = e^{At}h(0) + \int_{0}^{t} e^{A(t-\tau)}Bu(\tau)d\tau,
\end{equation}
where \( h(t) \) represents the state at time \( t \) and \( h(0) \) denotes the initial state.

\subsection{Discrete Parameter Solution}
For discrete-time analysis~\citep{xi2023quantifying}, we introduce the discrete time points with \( \mathcal{T} = \{t_k = k\Delta\} \), where \( \Delta = t_k - t_{k-1} \) represents a constant step size. For any time-dependent function \( f(t) \), we define its discretization as:
\[
f(k) := f(k\Delta) = f(t_k).
\]

To discretize the exact solution formula \eqref{eq:6.3} over the interval \( [t_k, t_{k+1}] \), we consider:
\begin{align*}
h(k+1) &= e^{A\Delta}h(k) + \int_{t_k}^{t_{k+1}} e^{A(\tau-t_k)}Bu(\tau)d\tau \\
        &= e^{A\Delta}h(k) + \int_{0}^{\Delta} e^{A\tau}Bu(\tau+t_k)d\tau.
\end{align*}

Assuming \( u(\cdot) \) remains constant over the interval \( [t_k, t_{k+1}] \), the equation simplifies to:
\[
h(k+1) = e^{A\Delta}h(k) + \left(\int_{0}^{\Delta} e^{A\tau}d\tau\right) Bu(t_k) = e^{A\Delta}h(k) + \Phi^{-1}(e^{A\Delta}-I)B\Delta u(k),
\]
where \( I \) denotes the identity matrix of appropriate dimensions, and \( \Phi \) is a shorthand notation for \( A\Delta \). Thus, the discrete-time state space model is:
\begin{align}
h(k+1) &= \mathcal{A}h(k) + \mathcal{B}u(k), \label{eq:6.4} \\
y(k+1) &= Ch(k+1), \label{eq:6.5}
\end{align}
where \( \mathcal{A} = e^{A\Delta} \) is the discrete state transition matrix and \( \mathcal{B} = \Phi^{-1}(e^{A\Delta}-I)B\Delta \) is the discrete input matrix.


\bibliographystyle{elsarticle-harv}
\bibliography{example}

\end{document}